\pdfoutput=1
\documentclass[11pt]{article}

\usepackage[]{emnlp2021}

\usepackage{times}
\usepackage{latexsym}

\usepackage[T1]{fontenc}
\usepackage[utf8]{inputenc}
\usepackage{csquotes}

\usepackage{microtype}

\usepackage{inconsolata}

\usepackage{amsmath}
\usepackage{booktabs}
\usepackage{graphics}
\usepackage{graphicx, multirow}
\usepackage{amsmath}
\usepackage[linesnumbered,ruled]{algorithm2e}
\usepackage{makecell}
\usepackage{array}
\usepackage{arydshln} 

\usepackage{enumitem}

\definecolor{deeppink}{rgb}{1.0, 0.08, 0.58}

\newcommand{\ourdataset}{Adv-CSFB}

\newcommand{\specialcell}[2][c]{%
\begin{tabular}[#1]{@{}l@{}}#2\end{tabular}
}

\title{ 
Clever Hans or Neural Theory of Mind?\\ 
Stress Testing Social Reasoning in Large Language Models}

\author{Natalie Shapira$^1$~~Mosh Levy*$^1$~~Seyed Hossein Alavi*$^{2,3}$~~Xuhui Zhou*$^4$\\
\textbf{Yejin Choi}$^{5,6}$~~\textbf{Yoav Goldberg}$^{1,5}$~~\textbf{Maarten Sap}$^{4,5}$~~\textbf{Vered Shwartz}$^{2,3}$ \\
$^1$ Bar-Ilan University~~ 
$^2$ University of British Columbia\\
$^3$ Vector Institute for AI~~$^4$ Carnegie Mellon University\\
$^5$ Allen Institute for Artificial Intelligence~~
$^6$ University of Washington\\
{\tt nd1234@gmail.com}}

\begin{document}
\maketitle

\begin{abstract}
The escalating debate on AI's capabilities warrants developing reliable metrics to assess machine ``intelligence.'' Recently, many anecdotal examples were used to suggest that newer large language models (LLMs) like ChatGPT and GPT-4 exhibit Neural Theory-of-Mind (N-ToM); however, prior work reached conflicting conclusions regarding those abilities. We investigate the extent of LLMs' N-ToM through an extensive evaluation on 6 tasks and find that while LLMs exhibit certain N-ToM abilities, this behavior is far from being robust. We further examine the factors impacting performance on N-ToM tasks and discover that LLMs struggle with adversarial examples, indicating reliance on shallow heuristics rather than robust ToM abilities. We caution against drawing conclusions from anecdotal examples, limited benchmark testing, and using human-designed psychological tests to evaluate models. \renewcommand{\thefootnote}{\empty}\footnotetext{* These authors contributed equally to this work.}
\end{abstract} 

\section{Introduction}
\label{sec:introduction}

Theory of Mind (ToM) is the ability to understand that other people have thoughts, beliefs, and emotions that differ from one's own \cite{wimmer1983beliefs}. As ToM is inherently linked to human cognition, imbuing machines with capabilities that mimic or resemble ToM has the potential to lead to the ``ELIZA effect'' \cite{weizenbaum1976computer}, wherein human-like intelligence or even sentience and consciousness is incorrectly ascribed to the machine \cite[e.g.,][]{kosinski2023theory,bubeck2023sparks}. 

In light of these possibly illusory ToM abilities, there is a pressing need to develop robust metrics for assessing Neural-ToM (N-ToM) in machines. This is particularly crucial given the escalating stakes of the debate on the extent to which machines possess ToM-like abilities and the potential ramifications of overblown claims in AI.\footnote{\url{https://futureoflife.org/open-letter/pause-giant-ai-experiments/}\\\url{https://amcs-community.org/open-letters/}}

Two recent papers addressed whether Large Language Models \cite[LLMs;][]{brown2020language, bommasani2021opportunities, Zhao2023survey} have a ToM, and came to opposite conclusions:  \citet{sap2022neural} shows they lack this ability and \citet{kosinski2023theory} claims this ability has emerged in the newer models spontaneously. The latter was criticized for its flawed methodology \cite{Marcus2023howNotTo}. \newcite{ullman2023large} further showed that simple changes to the ToM questions break LLMs. But to paraphrase the saying,  hype gets halfway around the world before rigorous experiments put on their boots; other researchers continue to spread the word about N-ToM, claiming that GPT-4 \textit{``has a very advanced level of theory of mind''} based on a few anecdotal examples \cite{bubeck2023sparks}. 

\paragraph{Do LLMs have robust N-ToM?} This paper aims to address the discrepancy and  limited scope of previous work (that each tested 2 tasks) by performing an extensive evaluation on 6 tasks targeting various aspects of ToM. We also experiment with different probing methods (i.e., generative QA format vs. probability of answer choices). We find that contemporary LLMs demonstrate certain N-ToM abilities, but these abilities are not robust (\S\ref{sec:experiments}).

\paragraph{ToM or Spurious Correlations?} We investigate through a series of experiments the factors influencing performance on N-ToM tasks. We show that LLMs perform worse on datasets that were designed to prevent annotation artifacts. We also enhanced the dataset originally proposed by \newcite{kosinski2023theory} to incorporate adversarial examples inspired by \newcite{ullman2023large}. We find that the performance of LLMs decreases for adversarial examples, suggesting that LLMs don't have robust ToM abilities but rather rely on shallow heuristics (\S\ref{sec:clever_hans}). 

We summarize these findings and additional insights in \S\ref{sec:discussion}. In particular, we warn against drawing conclusions from anecdotal examples, testing on a few benchmarks, and using psychological tests designed for humans to test models.\footnote{The code and data is available at: \url{https://github.com/salavi/Clever\_Hans\_or\_N-ToM}}
\section{Background: ToM and Clinical Tests}
\label{sec:theory_and_clinical_tests}
ToM has a long history starting in philosophy \cite{lewis1966argument} and later in psychology and cognitive science \cite{premack1978does}. ToM involves understanding mental states, beliefs, desires, intentions, and emotions of the self and of others. Clinical psychology tests were developed to test ToM abilities in humans, such as the false belief and faux pas tests detailed here.\footnote{For a detailed review see  \cite{osterhaus2022looking}.} 

\subsection{False Belief Test}
\label{sec:false_belief}
In a false belief test \cite{wimmer1983beliefs} the examinee is told a story in which a character in the story is exposed to partial information and therefore mistakenly believes in something that is not true (``false belief'') in contrast to the listener who is exposed to the full story.

A widely used clinical psychology task to assess false belief understanding is the \textit{Sally–Anne Test} \cite{baron1985does} or  \textit{unexpected transfer}.
In this test, Sally has a basket, and Anne has a box. Sally puts a marble in her basket and leaves the room. Anne takes the marble out of the basket and puts it in her box. The examinee is asked about \textbf{first order belief}, i.e. \textit{where will Sally look for her marble?}; about the \textbf{reality}, i.e. \textit{where is the marble?}; and about their \textbf{memory}, i.e. \textit{where was the marble in the beginning?}. 

The answers are that Sally will look in the basket, where she left the marble. Sally's belief is false because she is unaware of the marble's relocation to the box. However, a listener exposed to the entire story knows that the marble is no longer in Sally's basket and that Sally will look in the wrong place. 

In more complex versions, \textbf{Second Order Belief} question would be, \textit{where does Anne think Sally will look for her marble?}

In a different version of a false belief task, known as the \textit{Smarties Test} \cite{perner1987three}, the protagonist is dealing with \textit{unexpected content}, i.e., unaware of the actual contents of a container because of false labeling.

\subsection{Faux Pas Test} 
\label{sec:faux_pas}

Faux Pas occurs when \textit{``a speaker says something without considering if it is something that the listener might not want to hear or know, and which typically has negative consequences that the speaker never intended''} \cite{baron1999recognition}. An example of a faux pas situation is when a guest tells their hosts that they ``like cakes except for apple pie'', without realizing that the hosts have made an apple pie for them. 
The complexity of the situation depends not only on the content of the statement (``except for apple pie'') but also on the context in which it was made (e.g., the host had made an apple pie and the guest was unaware). Faux pas is the ``uh-oh!'' emotion most people would feel when they reveal the reality of the context. In this context, the statement wouldn't be problematic if the hosts made a cheesecake instead.

In the original test, the subject is told 10 stories that contain faux pas. At the end of each story, the subject is asked 4 questions: \textbf{detection} - \textit{In the story did someone say something that they should not have said?}; \textbf{identification} - \textit{What did they say that they should not have said?}; And two questions that differ by story:
\textbf{comprehensive} - e.g., Where does the event take place?,  and \textbf{false belief} - did they know or remember that?

\subsection{From Human Tests to Machine Tests}
Studies have explored the use of NLP techniques to model basic ToM skills. For example, in detecting mental states and emotions \cite{tausczik2010psychological, guntuku2017detecting, gordon2017formal,rashkin2018modeling,rashkin2018event2mind, shapira2021hebrew} or by generating a humorous response when the interlocutor is in a playful mood \cite{Shani2022, shapira2023evaluating}. Recent work is focused around creating datasets testing whether and to what extent models have ToM (see \S\ref{sec:data}). \textbf{It is important to note that the consequences of the success of these tests do not straightforwardly transfer from humans to models} (see \S{\ref{sec:discussion}}). 
\section{Data}
\label{sec:data}

\begin{table*}[t]
\centering
\resizebox{\linewidth}{!}{
\begin{tabular}{p{1.8cm}p{4.155cm}p{1cm}p{1.5cm}p{14cm}}
\hline
\textbf{Dataset} & \textbf{Inspired by Theory/Test} & \textbf{Test Size} & \textbf{Construc -tion} & \textbf{Example}\\ 
\toprule
\textbf{Triangle COPA} \citet{gordon2016commonsense} & Interpreting the social behaviour of fictional characters
& 100 & Experts & 
A circle is in the house moving around. A triangle bursts in through the door. The circle turns around and freezes. How does the circle feel? 
\newline \textbf{(a) The circle is surprised by the triangle's sudden entrance into the room.}
\newline (b) The circle is excited to see the triangle.  \\
\midrule
\textbf{SocialIQa} \citet{sap2019socialiqa} & Reasoning about motivations, what happens next and emotional reaction 
& 400 \newline random \newline sample \newline out of \newline 37,588 & Crowd- sourcing &  In the school play, Robin played a hero in the struggle to the death with the angry villain. \newline How would others feel afterwards? 
\newline (a) sorry for the villain
\newline \textbf{(b) hopeful that Robin will succeed}
\newline (c) like Robin should lose \\ 
\midrule
\textbf{ToMi} \citet{le2019revisiting}  & Unexpected transfer task, first and second order false belief; \cite{baron1985does} & 400 \newline random \newline sample \newline out of \newline above \newline 1000 & Synthetic &  Jackson entered the hall. Chloe entered the hall. The boots is in the bathtub. Jackson exited the hall. Jackson entered the dining\_room. Chloe moved the boots to the pantry. \newline(Memory) Where was the boots at the beginning?\textit{ (bathtub)} \newline(Reality) Where is the boots really? \textit{(pantry)} \newline(First order) Where will Chloe look for the boots? \textit{(pantry)} \newline(Second order) 7 Where does Chloe think that Jackson searches for the boots?	\textit{(bathtub)} \\
\hdashline
\textbf{ToMi'} \newline \textcolor{red}{This paper}, based on ToMi adjustments&  & 180 \newline questions \newline 30 \newline stories  & Experts &  <Same story as in ToMi> \newline(Memory) At the beginning, the boots were in the \textit{(bathtub)} \newline(Reality) The boots are really in the \textit{(pantry)} \newline(First order) Chloe will look for the boots in the  \textit{(pantry)} \newline(Second order) Chloe thinks that Jackson searches for the boots in the \textit{(bathtub)}  \\
\midrule
\textbf{epistemic \_reasoning} \citet{cohen2021exploring} &  Verbs, factive and non-factive, that describe epistemic mental states; intra-personal, inter-personal and inference reasoning; 
\cite{wimmer1983beliefs, hintikka1962knowledge}  
& 2000 & Experts with 10 templates  &  Premise: John knows that Ann thinks that there is milk in the kitchen. 
\newline Hypothesis: Ann thinks that there is milk in the kitchen. \textit{(Entailment = 1)} \newline
Hypothesis: John thinks that there is milk in the kitchen. \textit{(Entailment = 0)}
\newline\newline
Premise: John thinks that Ann knows that there is milk in the kitchen. 
\newline Hypothesis: Ann thinks that there is milk in the kitchen. \textit{(Entailment = 0)} \newline
Hypothesis: John thinks that there is milk in the kitchen. \textit{(Entailment = 1)}
\\
\midrule
\textbf{\ourdataset} \textcolor{red}{This paper}, based on \citeauthor{kosinski2023theory} \& \newcite{ullman2023large} & Unexpected content or transfer task, integrate commonsense reasoning, first-order false belief;   \cite{baron1985does, perner1987three} &  183 \newline questions \newline 40 \newline stories & Experts &  On the shelf, there is a bottle. It is full of beer and the label on this bottle says ``beer''. Mark walks into the room looking for beer and notices the bottle. He has never seen it before. He reads the label.
\newline (a) He opens the bottle and looks inside, He can clearly see that it is full of \textit{(beer) }
\newline (b) He believes that it is full of \textit{(beer)}
\newline (c) He calls his friend to tell them that he has just found a bottle full of \textit{(beer) }
\\  
\midrule
\textbf{FauxPas -EAI} \citet{shapira2023how} & Recognition of faux pas \cite{baron1999recognition} & 176 \newline questions \newline 44 \newline stories   & Experts \textcolor{white}{\_\_}and AI+Experts & 
Jeff was in an interview. When he finished the interview he sank into a couch in the lobby. Sarah and Tim, the executives who interviewed him, went out of the room and into the lobby while Sarah said: ``He asked for 179K, could have asked for much more. We need to make him an offer quickly at the salary he asked for''. Tim saw Jeff and said ``Oh, goodbye''.
\newline (Faux Pas) In the story did someone say something that they should not have said? \textit{(Yes)}
\newline (Identification) What did they say that they should not have said? \textit{(``He asked for..)}
\newline (Comprehensive) Who was interviewed? \textit{(Jeff)}
\newline (False Belief) Did Sarah know Jeff is sitting on one of the couches in the lobby? \textit{(No)}
\\ 
\bottomrule
\end{tabular}
}
\caption{Theory of Mind (ToM) datasets used in this work.}
\label{table:datasets}
\end{table*}

We used all datasets listed in Table~\ref{table:datasets} in our experiments. 
Below is a brief description of each dataset.
The creation of \textbf{ToMi'} which is based on ToMi is described immediately after the description of ToMi (\S{\ref{sec:tomi_adjustments}}).
The creation of \textbf{\ourdataset} (\S{\ref{sec:creation_Adv-CSFB}}) contains a description of the datasets it is based on.

\subsection{Existing Benchmarks \& Variants}
\paragraph{Triangle COPA \cite{gordon2016commonsense}.}
A set of 100 problems, each describes a short sequence of events involving the characters of the \newcite{heider1944experimental} film: two triangles and a circle moving around a box with a hinged opening. The questions require understanding the action sequence and social reasoning, and two answer choices are given.

\paragraph{SocialIQa \cite{sap2019socialiqa}.} A large-scale (38k) dataset for commonsense reasoning about social situations. Questions in SocialIQa require reasoning about people's motivations and mental states, causes and effects. The questions in SocialIQa were crowdsourced along with correct and incorrect answers. Additional distractors were added by using the correct answer for a different question on the same context, using a framework that mitigates stylistic artifacts. 
 
\paragraph{ToMi \cite{le2019revisiting}.}
Inspired by the \textit{Sally-Anne} test, ToMi is an improved iteration of prior datasets \cite{weston2015towards, grant2017can,nematzadeh2018evaluating}, comprising over 1,000 distinct stories and questions regarding memory, reality, and first and second-order false belief. This synthetic dataset was automatically generated for a range of essential objects and actions and was further processed for artifact prevention.\footnote{See Appendix~\ref{sec:chatgpt_failure} for an example.} 

\paragraph{ToMi Adjustments (ToMi')}
\label{sec:tomi_adjustments}
ToMi stories are in question-answering format. We randomly sampled 30 stories (each story has 6 questions, 180 questions in total) from the ToMi dataset and modified them to match a sentence completion format with the same meaning.\footnote{This was done manually by one of the authors.}  For example the question: \textit{``Where does Oliver think that Emma searches for the grapes?''}. Was adjusted to the following sentence completion task: \textit{``Oliver thinks that Emma searches for the grapes in the''}.

\begin{table*}[t]
    \centering
    \resizebox{\linewidth}{!}{
    \small
    \begin{tabular}{ll} 
        \toprule
         On the shelf, there is a \textcolor{blue}{bottle}.  & On the shelf \textcolor{red}{in the company's headquarters}, there is a \textcolor{red}{hard drive} \\
         \textcolor{blue}{It is full of} \textbf{beer}; \textcolor{blue}{there is} no \textbf{wine} \textcolor{blue}{in it}.  & \textcolor{red}{that contains only} \textbf{audio files} \textcolor{red}{and} no \textbf{video files.} \\
         Yet, \textcolor{blue}{the} label \textcolor{blue}{on this bottle says} `\textbf{wine}' and not `\textbf{beer}'. & Yet, \textcolor{red}{confusingly, its} label \textcolor{red}{clearly states} `\textbf{video files}' and not `\textbf{audio files}.' \\
         \textcolor{blue}{Mark walks into the room and notices the bottle.} &  \textcolor{red}{The newly hired computer engineer finds the hard drive on the shelf.} \\
         He has never seen \textcolor{blue}{it} before. He reads the label. & She has never seen \textcolor{red}{this hard drive} before. She reads the label.\\
         \bottomrule
    \end{tabular}
    }
    \caption{Variations that demonstrate the pattern similarity. Besides the lexical match (black) there are also semantic, grammatical, and pragmatic matches e.g., ``beer'' and ``audio files'' both share the same POS-tag and place in the parsing tree; ``full of'' and ``contains'' share the same semantic meaning for the purpose of the question.}
    \label{tab:example_diversity}
\end{table*}

\paragraph{Epistemic Reasoning \cite{cohen2021exploring}.}
  This dataset is part of BIG bench  \cite{srivastava2022beyond}. It combines ToM with natural language inference. The tests pertain to epistemic mental states \cite{wimmer1983beliefs} and epistemic logic \cite{hintikka1962knowledge}. This is done by using specific verbs related to knowledge and belief: factive (i.e., know, understand, recognize, see, remember, learn), and non-factive (i.e., believe, think, suspect, assume). The dataset contains 3 types of tests: (1) \textbf{intra-personal tests}:  reasoning about the mental states of a single agent; (2) \textbf{inter-personal tests}:  reasoning about the mental states of multiple agents; and (3) \textbf{inference reasoning}: recognizing that other agents are making inferences (i.e., if X entails Y, and Bob believes that X, then, it is reasonable to conclude that Bob believes Y).

\paragraph{FauxPas-EAI \cite{shapira2023how}.}
Based on the clinical faux pas test \cite{baron1999recognition}, the set contains 44 stories (22 faux pas and 22 equivalent control) with 4 corresponding questions. The stories require both social reasoning skills and detecting false belief. The stories were created by experts and a small part of the stories was created by ChatGPT with rephrasing and fixes by experts.

\subsection{Creation of \ourdataset}
\label{sec:creation_Adv-CSFB}

Inspired by the disagreeing conclusions reached by prior work, we introduce the \textit{ADVersarial CommonSense with False-Belief} (\ourdataset) dataset. \ourdataset{} contains 110 examples of the unexpected contents task and 73 examples of the unexpected transfer task (\S\ref{sec:false_belief}). Each manually-created example in the dataset consists of a short paragraph describing two objects $O_1$ and $O_2$, and is followed by questions pertaining to \textbf{reality}, i.e. whether a certain container contains $O_1$ or $O_2$, and the protagonist's \textbf{belief} regarding the content. 

% The three types of examples in this dataset 
The examples in \ourdataset{} are categorized to \emph{false belief}, i.e. the original examples from ToM-k \cite{kosinski2023theory}, \emph{true belief}, and \emph{adversarial} examples inspired by \newcite{ullman2023large}. 

\paragraph{False Belief.} In the false-belief examples from \newcite{kosinski2023theory}, the 
protagonist's belief about the content of the container is different from its actual contents. The examples are variants of the corresponding original tests from psychology, e.g. the unexpected contents examples are variants of the Sally-Anne test.
Notably, \citeauthor{kosinski2023theory} only created false-belief scenarios.

\paragraph{True Belief.} For a more fair evaluation setup, we enhance the unexpected contents task with \emph{true belief} examples, i.e. in which the protagonist's belief about the content of the container is the same as its actual contents. We do so by modifying each of the \emph{false belief} examples such that the label now indicates the true content of the container, $O_1$. We mention the alternative content $O_2$ in a way that doesn't change the answer, e.g. Mark walks into the room looking for $O_2$ but finds a bag with $O_1$ labelled as ``$O_1$''. One author of this paper created a variation for each applicable example, which was then verified by another author. 

\paragraph{Adversarial Examples.} \newcite{ullman2023large} showed that LLMs that achieve near-perfect performance on the \emph{false belief} examples fail to solve a number of adversarial examples where new information is introduced. In particular, LLMs still predict false belief even when new information suggests that the protagonist should know the truth. For example, the LLM predicts that a protagonist looking at a bag full of popcorn that is labelled as ``chocolate'' believes the bag is full of chocolate, even if the bag is transparent or if the protagonist cannot read. \citeauthor{ullman2023large}'s counter examples are sufficient in showing that LLMs did not robustly acquire ToM abilities. To further quantify the LLMs' abilities, we created up to 4 additional examples for each of the \emph{false belief} examples, following each of the alterations suggested by \newcite{ullman2023large}: transparent access, uninformative label, trustworthy testimony, and late labels for the unexpected contents task, and transparent access, in$\rightarrow$on, trustworthy testimony, and other person for the unexpected transfer task (see Appendix~\ref{sec:ullman_variations} for an example for each variation). Again, the examples were created by one author and verified by another.

\section{Experiments \& Results}
\label{sec:experiments}

To investigate the ToM abilities of LLMs, we designed experiments that explore various aspects. The first experiment presents a meta-evaluation of 15 LLMs evaluated on multiple ToM-related datasets in a zero-shot manner (\S\ref{sec:meta_eval_tom}). We then investigate to what extent LLMs are sensitive to the probing method (\S\ref{sec:probing_method}). 

\paragraph{LLMs} We examine the performance of 15 different LLMs of different sizes:
FlanT5: \texttt{\small flan-t5-\{small, base, large, xl, xxl\}} \cite{chung2022scaling}, FlanUl2 \cite{tay2022unifying}, GPT-3 (\texttt{\small text-davinci-002, text-davinci-003}), GPT-3.5 / ChatGPT (\texttt{\small gpt-3.5-turbo-0301}), GPT-4 (\texttt{\small gpt-4-0314}) \cite{brown2020language,Ouyang2022InstructGPT}, 
and Jurassic2: \texttt{\small j2-\{jumbo-instruct, grande-instruct, jumbo,} \texttt{\small grande, large\}}.\footnote{\tiny\url{https://www.ai21.com/blog/introducing-j2}} We provide technical details regarding prompting and decoding parameters in Appendix~\ref{app:llm}. 

\subsection{How well do LLMs perform on ToM tasks? Meta-Evaluation}
\label{sec:meta_eval_tom}

\begin{figure}[t]
    \centering
    \includegraphics[width=\columnwidth]{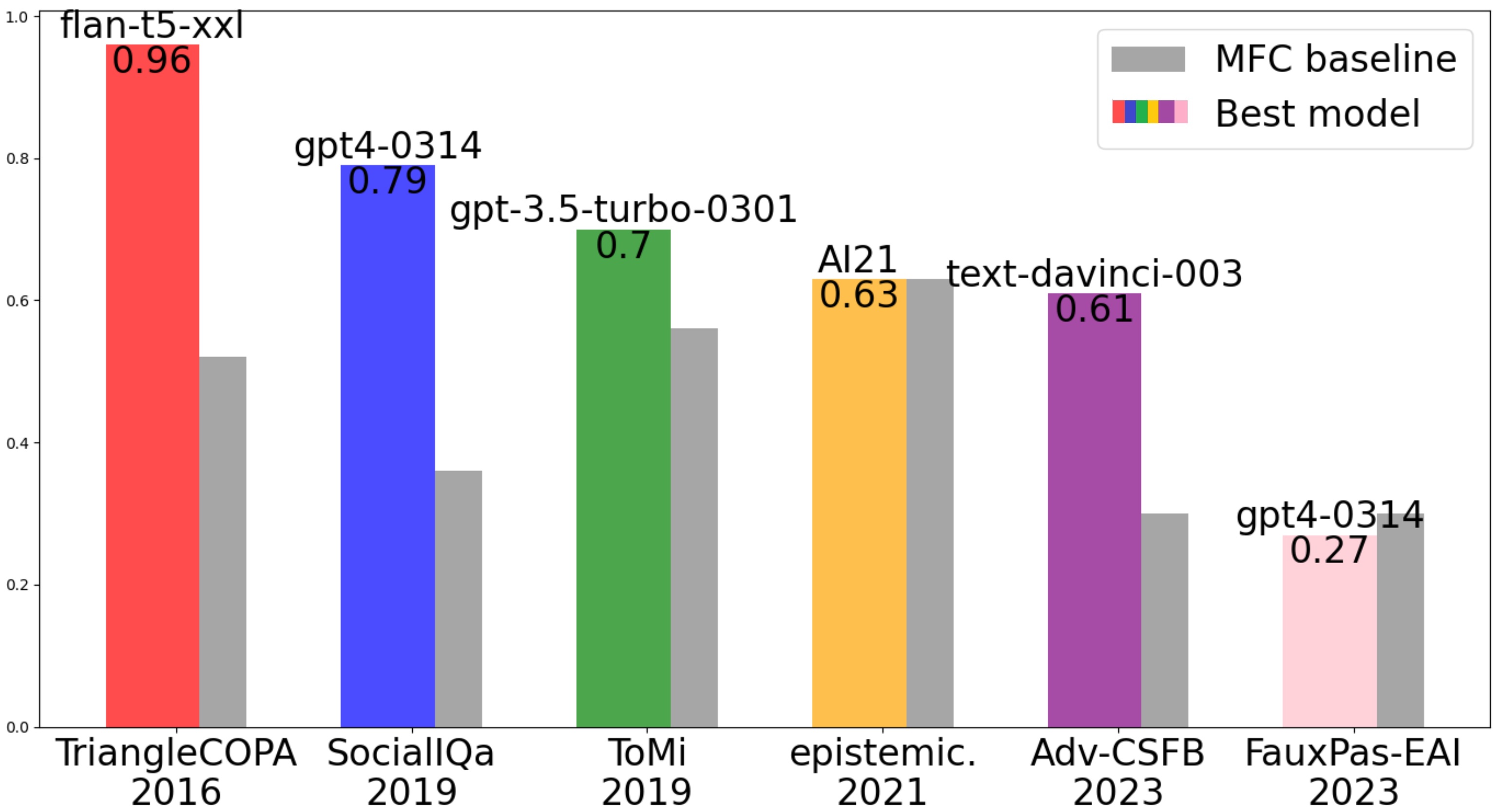}
    \caption{Accuracies of top-performing models on each of the ToM tasks, compared to a most frequent class (MFC) baseline. For several datasets, the best model achieves performance comparable to the MFC baseline, suggesting very limited ToM ability. 
    }   
    \label{fig_datasets_leader_bord}

\end{figure}
We conducted an evaluation of the performance of 15 LLMs in a zero-shot manner \cite{liu2021pre} on all ToM-related datasets considered (\S{\ref{sec:data}}), and compare to a most-frequent-class (MFC) baseline that always predicts the most frequent answer in each dataset. The summary of the results is presented in Figure \ref{fig_datasets_leader_bord}, and the complete results in Appendix~\ref{sec:all_results}. 

Our findings demonstrate that while some LLMs achieve near perfect accuracies on some datasets  (e.g., TriangleCOPA with 96\% accuracy by \texttt{flan-t5-xxl}), others datasets remain challenging for LLMs with considerably lower performance. 
For instance, the best performing LLM on the FauxPasEAI datasets is inferior to a simple most-frequent-class baseline, 
indicating the difficulty level of these datasets.

Notably, the best LLMs performance seems correlated to the dataset's age
(i.e., the older the dataset, the better the performance). This trend could be attributed to the fact that the increasing sophistication of LLMs is driving the creation of more challenging datasets, prompting researchers to set a higher bar. Another possibility is that LLMs have had more opportunities to train on the older datasets, resulting in better performance (see \S\ref{sec:data_contamination}).

Based on this meta-evaluation, our results suggest are that while some models exhibit strong ToM abilities on some datasets, \textbf{no model robustly exhibits ToM on all datasets}. These findings are consistent with  \newcite{sap2022neural} and  \newcite{ullman2023large}.

\begin{table}[t]
  \centering
  \small
    \begin{tabular}{llccc}
    \toprule
    & & \textbf{LM} & \textbf{MC} & \textbf{CoT} \\
    \midrule
    \multirow{3}{*}{\rotatebox{90}{Siqa}}
            & davinci-003 & 55 & 60 & 68 \\
            & GPT-3.5 & - & 67 & 69 \\
            & GPT-4 & - & 79 & 72 \\                 
    \midrule
    \multirow{3}{*}{\rotatebox{90}{ToMi}} 
            & davinci-003 &  67 & 67 & 71 \\
            & GPT-3.5 & - & 70 & 73 \\
            & GPT-4 & - & 70 & 73 \\
    \bottomrule
    \end{tabular}%
    \caption{Accuracy of the recent GPT models on a random sample of 400 instances from SocialIQa (Siqa) and ToMi. The probing method affects the performance. For example, in Siqa there is a 7\% difference in the accuracy of GPT-4 between MC-probing and CoT-probing.}
  \label{tab:extract_method}
\end{table}%

\subsection{How sensitive are LLMs to the probing technique?}
\label{sec:probing_method} 

We examine the effect of the different probing methods detailed below on LLM performance. Certain techniques have shown to be superior to others \cite[e.g.,][]{wei2023chainofthought}. However, we argue that to claim that a model has N-ToM abilities, it is essential that it performs well across probing techniques. On one hand, the most efficient method can potentially reveal latent capabilities, while on the other hand, there is a reasonable expectation for LLMs to succeed in the tasks regardless of the probing approach used to extract information. 

\paragraph{LM-probing} predicts the option with the highest probability  \cite{brown2020language,sap2022neural}.

\paragraph{MC-probing} prompts the LLM with the context, question, and answer choices, and asks it to generate the answer in the form of ``a, b, c''. This method is applicable for LLMs such as GPT-3.5 and GPT-4 that don't produce probabilities \cite{hu_fine-grained_2022}.

\paragraph{CoT-probing} asks the model to first ``reason'' about the question step-by-step and then give a final answer, which generally contributes to better performance \citep{wei2023chainofthought}.\footnote{We use the zero-shot setup without providing any reasoning examples.} 

Table~\ref{tab:extract_method} shows that the probing techniques influence the LLM performance on both datasets. CoT generally demonstrates enhanced performance, as supported by prior research \cite{camburu2018snli,shwartz-etal-2020-unsupervised, wei2023chainofthought}. Nonetheless, there are cases where this trend does not hold, since the reasoning may occasionally result in erroneous conclusions \cite{jung-etal-2022-maieutic}. 
\section{Clever Hans vs. Generalized Reasoning}
\label{sec:clever_hans}

We conducted a series of experiments aimed to enhance our understanding of the factors influencing performance in the context of N-ToM tasks. The research question that guided us was: Do the models that solve the tasks possess a general ability or do they rely on memorization and shallow heuristics \cite[``Clever Hans'';][]{kavumba-etal-2019-choosing}? 
We detail the experiments and findings below.

\subsection{Do LLMs Rely on Spurious Correlations?}
\label{sec:experiment_negative_examples}
\begin{table}[t]
\centering
\scriptsize
\begin{tabular}{@{}l|cc|c@{}}
\toprule
\multicolumn{1}{c|}{\textbf{Dataset}} & \multicolumn{2}{c|}{\textbf{ToMi'}} & \textbf{ToM-k} \\ \midrule
\multicolumn{1}{c|}{\textbf{Subset}}  & \textbf{All question} & \textbf{\begin{tabular}[c]{@{}c@{}}No second \\ order\end{tabular}} & \textbf{All questions} \\ \midrule
text-davinci-003 & 10 & 21 & 87 \\ 
GPT-3.5 & 27 & 48 & 65 \\
GPT-4 & 20 & 52 & 87 \\
\bottomrule
\end{tabular}
\caption{Comparison of LLMs' accuracy on ToM-k, which contains positive examples only, and on ToMi', which contains both positive and negative examples (manually adjusted from ToMi to be of the same probing type as ToM-k). ToM-k contains only first-order questions. The subset ``No second order'' was created manually to better compare to ToM-k dataset. Lower accuracy might suggest the dataset is more robust to spurious correlations.}
\label{tab:replication_table}
\end{table}

ToMi and ToM-k are datasets that examine the unexpected transfer false belief problem. While \mbox{ToM-k} contains only simple positive examples (variants of the original Sally-Annie test), ToMi also contains simple alternations such as omission or duplication of information that create negative examples (see example in Appendix~\ref{sec:chatgpt_failure}) and second-order questions.

To ensure a fair comparison between the question answering format of ToMi and the sentence completion format of ToM-k (see the effect of probing methods on performance in \S{\ref{sec:probing_method}}), we adjusted ToMi to match the sentence completion format (details about the adjustments can be found at \S{\ref{sec:tomi_adjustments}}). 
Additionally, we analyzed the results separately for second-order questions in order to facilitate a more accurate comparison with the \mbox{ToM-k} dataset. 

Table \ref{tab:replication_table}, shows significantly lower scores in ToMi'. The notable discrepancy between the performance of the two datasets suggests that the model's abilities are not based on generalization. Instead of true understanding of the problem at hand, such as accurately determining one's exact thoughts, the model might be recognizing patterns from the Sally-Anne story in other ToM-k examples and generating responses based on those patterns. Conversely, the performance on ToMi' is worse because it is more robust to spurious correlations.

\begin{figure*}[t]
    \centering
\includegraphics[width=\textwidth]{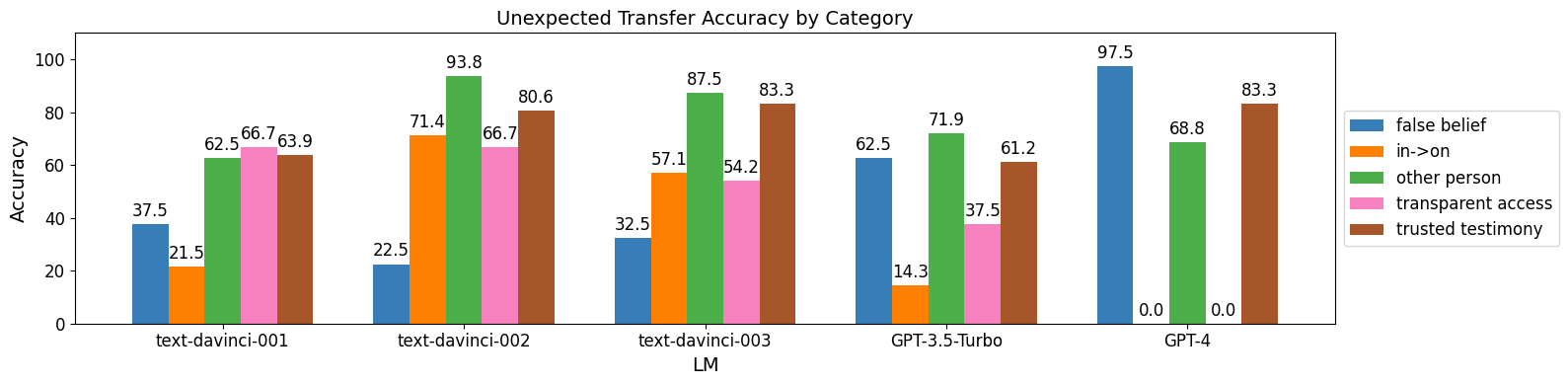}
    \caption{Performance of a range of GPT models on various categories within the unexpected transfer segment of \ourdataset{}. The results are the average accuracy of question 2 (e.g. \textit{Maria thinks that the bananas are in the \_}) and question 3 (e.g. \textit{When Maria comes back, she will first look for the bananas in the \_}), which specifically focus on an agent's beliefs rather than objective truth. Notably, GPT-4 achieves an accuracy of 97\% on the subset of false belief samples (the original examples from ToM-k), while failing on adversarial samples that involve transparent access or relationship change (in$\rightarrow$on).}
    \label{fig_unexpected_transfer}
\end{figure*}
\begin{figure*}[t]
    \centering
\includegraphics[width=\textwidth]{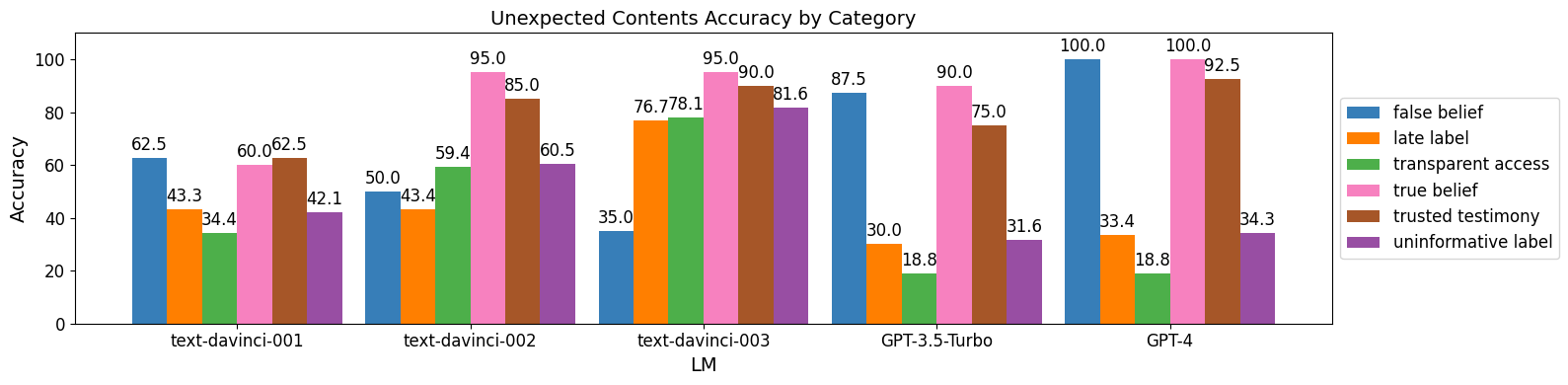}
    \caption{Performance of a range of GPT models on various categories within the unexpected content segment of \ourdataset{}. The results are the average accuracy of question 2 (e.g. \textit{He believes that it is full of \_}) and question 3 (e.g. \textit{He calls his friend to tell them that he has just found a bottle full of \_}), which specifically focus on an agent's beliefs rather than objective truth.}
    \label{fig_unexpected_contents}
\end{figure*}

\subsection{Is N-ToM Robust to Adversarial Changes?}
\label{sec:experiment_Adv-CSFB}

To test the robustness of the LLMs' N-ToM, we test the performance of GPT models on each of the categories in \ourdataset{} (\S\ref{sec:creation_Adv-CSFB}), using MC-probing. To ensure correct formatting and prevent unintended outputs (e.g., explanation of why the answer is correct), we prepend to the prompt \textit{one} out-of-domain example from ToMi, which has a similar format. We report the average accuracy of questions 2 and 3, both focusing on an agent's belief rather than objective truth. Finally, to ensure maximum reproducibility of the results, we set the temperature to 0. \textbf{Our main finding is that LLMs don't exhibit robust performance across different categories. In particular, later LLMs excel in some categories while completely failing on others.} We provide details below. 

Figure~\ref{fig_unexpected_transfer} illustrates the performance of a range of GPT models on different categories within the unexpected transfer segment of \ourdataset{}. It is evident that both \emph{false belief} (i.e. the original examples from ToM-k) and \emph{trusted testimony} (i.e., someone tells the protagonist that the object has been moved) have improved in newer models. GPT-4 achieves 97.5\% and 83.3\% on the two categories respectively. Nevertheless, there has been a gradual decline in the performance of subsequent models on other categories, such as \emph{other person} (from 93.8\% by \texttt{davinci-002} to 68.8\% by GPT-4), \emph{in$\rightarrow$on} (from 71.4\% by \texttt{davinci-002} to 0\% by GPT-4), and \emph{transparent access} (from 66.7\% by \texttt{davinci-002} to 0\% by GPT-4). 

Figure~\ref{fig_unexpected_contents} showcases the performance of the GPT family on various categories within the unexpected contents segment. It becomes apparent that, akin to the unexpected transfer segment, newer models such as GPT-3.5-Turbo and GPT-4 demonstrate improved performance in handling samples that involve \emph{false belief} and \emph{transparent access} (i.e., the container is transparent). Furthermore, nearly all models since \texttt{text-davinci-002} exhibit strong performance on \emph{true belief} samples. However, both GPT-3.5-Turbo and GPT-4 experience a substantial decline in performance compared to their earlier counterparts when it comes to \emph{transparent access}, \emph{late label} (e.g., the protagonist is the one who wrote the label), and \emph{uninformative label} (i.e., the protagonist can't read the label).

We regenerated the responses multiple times, consistently obtaining similar results, so we can conclude that the models exhibit confidence in their predictions, even if they are incorrect. It is important to note, however, that the results obtained from LM-probing may slightly differ from MC-probing. In MC-probing, even with our 1-shot setup, the model may produce responses that are not applicable, such as ``none of the above'' or ``both''. This is particularly noticeable in verbose models like GPT-3.5-Turbo and GPT-4. These models tend to be careful to avoid providing incorrect answers and, as a result, generate longer phrases. With that said, as we argue in \S\ref{sec:probing_method}, a LLM exhibiting robust N-ToM ability should be able to answer questions correctly regardless of the probing method. 

\subsection{Are Spurious Correlations a Trend?}
\label{sec:different_splits}

In the previous experiment \S{\ref{sec:experiment_Adv-CSFB}}, we saw that the datasets contain both difficult and easy questions. Here we show this recurring phenomenon across two ToM datasets, inspired by the analyses in \citet{sap2022neural}. 

\begin{figure}[t]
    \centering
\includegraphics[width=\columnwidth]{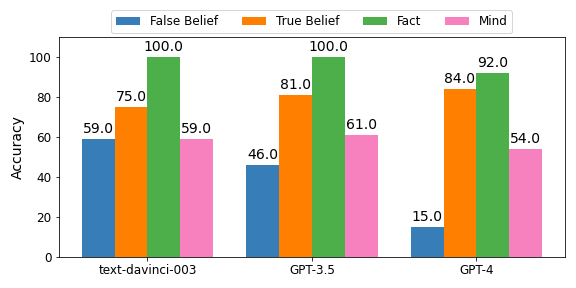}
    \caption{ToMi's accuracies with different splits of the dataset. While GPT-3.5 (the best-performing model) achieves a total of 0.7 accuracy score (see Figure~\ref{fig_datasets_leader_bord}), it achieves only 0.46 on the subset questions ``false belief''.}   
    \label{fig_splits_ToMi}
\end{figure}
Figure \ref{fig_splits_ToMi} describes ToMi accuracies on different question types; ToMi contains questions about facts vs. beliefs (mind), and specifically about true or false beliefs.
While GPT-3.5 (the best-performing model) achieves 81\% accuracy, on the subset questions ``false belief'', it achieves only 46\%, close to random performance.

\begin{figure}[t]
    \centering
    \includegraphics[width=\columnwidth]{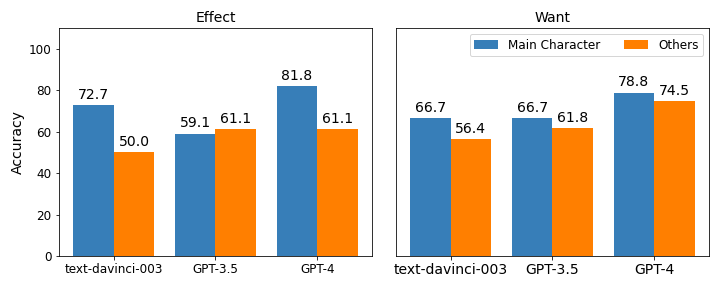}
    \caption{SocialIQa's accuracies for the questions that focus on the main character vs. others. While GPT-4 (the best-performing model) achieves a total of 0.79 accuracy score, it achieves only 0.61 on the subset questions of ``others effect''.}   

    \label{fig_splits_SocialIQa}

\end{figure}
Figure~\ref{fig_splits_SocialIQa} shows the SocialIQa accuracies for questions focusing on the main character vs. others. While GPT-4 (the best-performing model) achieves a total of 79\% accuracy score, on the subset questions of ``others'', it achieves only 74.5\%.

\section{Summary of Findings and Insights}
\label{sec:discussion}
We investigated whether modern LLMs robustly display N-ToM abilities. By quantifying their performance on 6 N-ToM benchmarks, we found that while some datasets have been nearly ``solved'' (e.g., TriangleCOPA with 96\% accuracy by \texttt{flan-t5-xxl}), others remain challenging for LLMs with considerably lower performance (e.g., FauxPas-EAI with 27\% accuracy by GPT-4, which is even below the majority baseline). We also created \ourdataset, a new ToM benchmark designed to uncover whether LLMs solve ToM questions for the right reasons, or merely rely on surface cues and shallow heuristics. 

\paragraph{So... Do LLMs have ToM?}
Our results show that while some datasets have been successfully solved, others remain challenging for LLMs. Thus, \textbf{models do not have \emph{robust} N-ToM abilities.}
These findings are inconsistent with \newcite{kosinski2023theory}, who claimed that ToM has emerged in LLMs as a byproduct of their development, a claim  further echoed by \citet{bubeck2023sparks}. We argue that these conclusions were over-generalized based on a specific aspect of ToM and a small number of examples (40 for \newcite{kosinski2023theory} and 10 for \citet{bubeck2023sparks}). Following \citet{ullman2023large}, we empirically showed that even the best models fail on small variations of the original tasks, proving that even GPT-4 does not display robust N-ToM abilities.

\paragraph{Clever Hans, Heuristics \& Shortcuts}
The performance gaps between different question types suggests that \textbf{LLMs rely on shortcuts, heuristics, and spurious correlations}, which often lead them astray. In \ourdataset{} (\S\ref{sec:experiment_Adv-CSFB}), the bad performance on some of the adversarial categories might be partly attributed to \emph{reporting bias} \cite{gordon2013reporting,shwartz-choi-2020-neural}. People don't share obvious facts \cite{grice1975logic}, so it is likely that LLMs are biased towards generating surprising rather than unsurprising continuations. In most of these categories, the protagonist belief is the same as the truth, making a boring story. 

Furthermore, the newer models such as GPT-3.5 and GPT-4 are trained in addition to the LM objective to follow natural language instructions and generate helpful answers. This might make them cooperative and lead to LLMs assuming that all details are important, rather than that the input is adversarial. For example, they might pay too much attention to the mention of the false label in the unexpected contents task, failing to see that the label doesn't matter if the person can't read it or if the container is transparent. The fact that LLMs perform reasonably well on true belief examples (Figure~\ref{fig_unexpected_contents}) might be attributed to recency bias \cite{OConnor2021-xj}, since the correct content is typically the last one to be mentioned.

Finally, we reassess the finding of \newcite{sap2022neural} that LLMs perform better on predicting the mental states of the main character vs. others (SIQA, \S\ref{sec:experiment_negative_examples}); \newcite{sap2022neural} suggested that this might be due to centering theory \cite{grosz-etal-1995-centering}, according to which texts tends to focus on describing a single protagonist. 

\paragraph{ELIZA Effect \& Anecdotal Generative vs. Automatic Large-Scale Multiple-Choice Testing}
The impressive anecdotal examples produced by LLMs in generative settings \cite[e.g., observed with ChatGPT and GPT4 web-demo;][]{bubeck2023sparks}, tends to captivate non-expert individuals. However, it is important to recognize that these models are specifically designed to generate text that appears high-quality to human observers \citep{Ouyang2022InstructGPT}.  
This inherent bias in their design can lead to the ``ELIZA effect'' \cite{weizenbaum1976computer,shapira2023how}, i.e. the human assumption that computer behaviors are analogous to human behaviors. Thus, \textbf{the illusion that a LLM has acquired human-like N-ToM often says more about the humans reading the text than about the model itself} \cite{Whang2023nytNeuralToM}.

Moreover, later models are by design trained to practice ``epistemic humility'' \cite[i.e., hedge and provide multiple possible answers;][p .17]{Ouyang2022InstructGPT}. This often leads them to provide rationales for each given answer without committing to actually answering the question. But humans might fall prey to confirmation bias and simply see the right answer and its rational and conclude that the model has gotten it correctly. We thus argue that in order to conclude whether a certain model possesses a certain ability, it is crucial to quantify the performance across multiple large-scale datasets, preferably using an automatic evaluation method. 

\paragraph{Using psychological tests designed for humans on LLMs}
In clinical psychology, tests designed for humans are carefully constructed and vetted to ensure that they have external and internal validity, i.e., that they are measuring what they aim to measure \cite{frank2023experimentology}. 
While there is evidence that a person's success in one ToM task can indicate their ToM abilities \cite[e.g.,][]{milligan2007language}, this does not necessarily transfer to models.  
Therefore, it is important to be cautious when drawing conclusions about ToM in models based on their performance on a few tasks \cite{Marcus2023howNotTo}. 
In general, when a system succeeds on an instrument designed for humans, we can't draw the same conclusions as we would for humans (e.g., that they have ToM). Instead, we need to consider other explanations (e.g., that they are relying on heuristics). The same holds in the other direction, when analyzing how models work in order to learn about the human brain. 

\paragraph{Machine intelligence and Anthropomorphism}
Relatedly, our results also point to a need for caution when discussing the abilities of machines in relation to concepts referring to human cognition, such as Theory of Mind.
While it is common in computer science to use human-related concepts and metaphors for AI systems, we caution readers to interpret ``\textit{neural} ToM'' carefully and without aiming to make claims about ``AI cognition,'' especially since given our propensity for anthropomorphizing non-human animals and computers \cite{epley2007seeing,kim2012anthropomorphism}; our measuring the performance on these benchmarks is not meant as an endorsement of the pursuit of a human-like social intelligence for AI systems.\footnote{We leave the question of whether LLMs, AI, or even any non-biological entity could develop human-like cognition and Theory of Mind up to philosophers.}
Instead, in light of the hype around AI and it's ``intelligence,'' we sought out to provide a more sober look at the empirical performance of LLMs on tasks related to social intelligence and ToM.

\paragraph{``Solving'' a ToM benchmark is necessary but not sufficient} 
Methodologically, if a model fails at least one ToM task, it does not have ToM in general. Success on one example or task is not a sound proof that a model has ToM. \textbf{Future work will need to continue to develop benchmarks testing various ToM aspects}, and these benchmarks will need to be designed to assess LLMs directly rather than using clinical tests designed for humans.  

Additionally, reporting the aggregated performance of LLMs on benchmarks obscures the performance differences across questions of different types and complexities.
To overcome this, one approach is to pair a difficult question with an easy question, requiring model to answer both correctly. This methodology resembles the ``joint score'' employed in FauxPas-EAI, \ourdataset, and ToMi. 
In situations where pairing is challenging, a recommendation for future works is that dataset difficulty could be evaluated by calculating the final score across different splits of the dataset. 
The difficulty level of the dataset can then be determined based on the lowest score obtained among these splits. 

\paragraph{Emergence vs. Supervised Learning vs Training on the Test set}
\label{sec:emergence}
Prior work claimed that ToM abilities emerged as a byproduct of the LLM training \cite{kosinski2023theory}. We argue that \textbf{claims about emergence are (i) unfounded, and (ii) unfalsifiable without access to the LLMs' training data}. To make a statement regarding emergent ToM, a careful experiment is needed to ensure that ToM did indeed appear spontaneously and not as a result of other factors such as training on related datasets, exposure to descriptions of clinical tests online, interactions with users, and more.\footnote{OpenAI acknowledged that GPT-4 was trained on test data from BIGBench \cite[][footnote 5]{OpenAI2023gpt4}.} However, since the data used to train the GPT models is not publicly available, it is impossible to quantify the degree of the potential data leakage.\footnote{See Appendix~\ref{sec:data_contamination} for an attempt to quantify such data leakage.}
We echo calls by \citet{dodge2021documentingC4} for increased transparency and open-access to the training data of LLMs, which is crucial for scientifically valid and reproducible experiments \cite{rodger2023closedModelsBlog}.

\paragraph{Improving neural ToM abilities (with CoT or other methods)}
Our objective in this study is not to measure benchmark performance or climb leaderboards. It is feasible that techniques such as chain-of-thought prompting \cite[CoT;][]{weichain} would enhance the performance of GPT-4 on tasks where it currently performs poorly. Nevertheless, we need to exercise caution to \textbf{ensure that the utilization of methods like CoT or others does not excessively guide the models} by essentially revealing the task structure to them---just like Clever Hans who appeared proficient in math merely due to subtle hints given by the owner.
 
\section{Conclusion}
\label{sec:conclusion}

Based on our research and replication studies, we conclude that contemporary LLMs demonstrate an enhanced yet limited degree of Theory of Mind abilities. We find that their ToM abilities are often not robust, and in some instances, we identify evidence of their over-reliance on simple heuristics rather than robust generalized reasoning.
\section*{Limitations} 

\paragraph{Benchmark scope and human ambiguity} The datasets used in this study were limited in scope and size; ToM is required in most human interaction, and thus unbounded in scope. In addition, parts of the datasets could be ambiguous, either due to lack of context or inherent ambiguity \cite{plank-2022-problem}. 
Due to this potential ambiguity, some LLMs were safeguarded and refused to answer certain questions; while we attempted to instruct them to respond in the correct format, some LLMs still did not output the right format. This was only an issue for MC-probing, but probability distributions were not available for all LLMs. 
Future work should investigate how to mitigate this issue via better instructions or methods that map generated answers to multiple choice better \cite[e.g.,][]{niu-etal-2021-semantic,bulian-etal-2022-tomayto}.

\paragraph{Limited text-only LLMs} Our experiments were conducted with a limited number of LLMs that were accessible at the time of writing, and we did not explore the full spectrum of LLMs that are currently available. Future work could explore the N-ToM abilities displayed by other LLMs, and additionally, explore multimodal models.

\section*{Ethical Statement}

\paragraph{Data.} All the existing and new datasets used in this study are publicly available. The narratives were evaluated by the authors to ensure that they do not contain offensive content.

\paragraph{Models.} LLMs may generate offensive content if prompted with certain inputs. However, we used them for evaluation only, with non-offensive inputs, and we did not record their responses.

\section*{Acknowledgements}
We would like to thank Uri Katz, Royi Rassin, Ori Shapira, Alon Jacoby, Rotem Dror, and Amir DN Cohen for helpful discussions. We thank OpenAI for access to their APIs including GPT-4, and AI21 for the generous budget for using their platform API.
This project was partially funded by the European Research Council (ERC) under the European Union's Horizon 2020 research and innovation program, grant agreement No. 802774 (iEXTRACT), the Computer Science Department of Bar-Ilan University, the Vector Institute for AI, Canada CIFAR AI Chairs program, an NSERC discovery grant, DARPA MCS program through NIWC Pacific (N66001-19-2-4031), and AI2. 

% Entries for the entire Anthology, followed by custom entries
\bibliography{custom,anthology}
\bibliographystyle{acl_natbib}

\section{Appendices}

\subsection{Example of ChatGPT Failure}
\label{sec:chatgpt_failure}

\begin{table*}[!ht]
    \centering
    \small
    \begin{tabular}{ll} 
        \toprule
        False Belief (Positive Example) & True Belief (Negative Example) \\ \toprule
Natalie and Michal are in the same room.                &  Natalie and Michal are in the same room. \\
Natalie puts a ring in \textbf{box A}.                           & Natalie puts a ring in box A. \\
\textcolor{red}{Natalie leaves the room.}               & Michal opens box A,  \\
Michal opens box A,                                     & takes the ring from inside and transfers it to \textbf{box B}.  \\
takes the ring from inside and transfers it to box B.   & Where will Natalie look for the ring? \\
\textcolor{red}{Natalie enters the room.}               & \\
Where will Natalie look for the ring?                   & \\
\toprule
 \texttt{ChatGPT:} & \texttt{ChatGPT:} \\
Natalie will look for the ring in \textbf{box A},                & Natalie will most likely look in \textbf{box A},  \\
where she initially put it.                             & where she initially put the ring. \\ 
She is not aware of                                     & Since Natalie did not witness \\
Michal's actions                                        &  Michal moving the ring to box B, \\ 
and assumes that the ring is still in the original box. & she would assume that the ring is still in box A. \\ 
Therefore, her natural inclination would be             & It is possible that Natalie may eventually check box B, \\ 
to look in the place where she left it.                 & but her initial instinct would be to look in box A. \\
         \bottomrule
    \end{tabular}
    \caption{An example of ChatGPT failure on an instance from ToMi.}
    \label{tab:example_positive_negative}
\end{table*}

Table~\ref{tab:example_positive_negative} shows an example from the Tomi dataset. The unexpected transfer test discusses an unexpected (false belief) rather than trivial (true belief) case. ChatGPT solves the more complex task (false belief) while failing on the trivial task, likely due to its exposure to the Salley-Anne task.

\subsection{\citeauthor{ullman2023large}'s Variations}
\label{sec:ullman_variations}

\begin{figure}[!ht]
    \centering
    \includegraphics[width=\columnwidth]{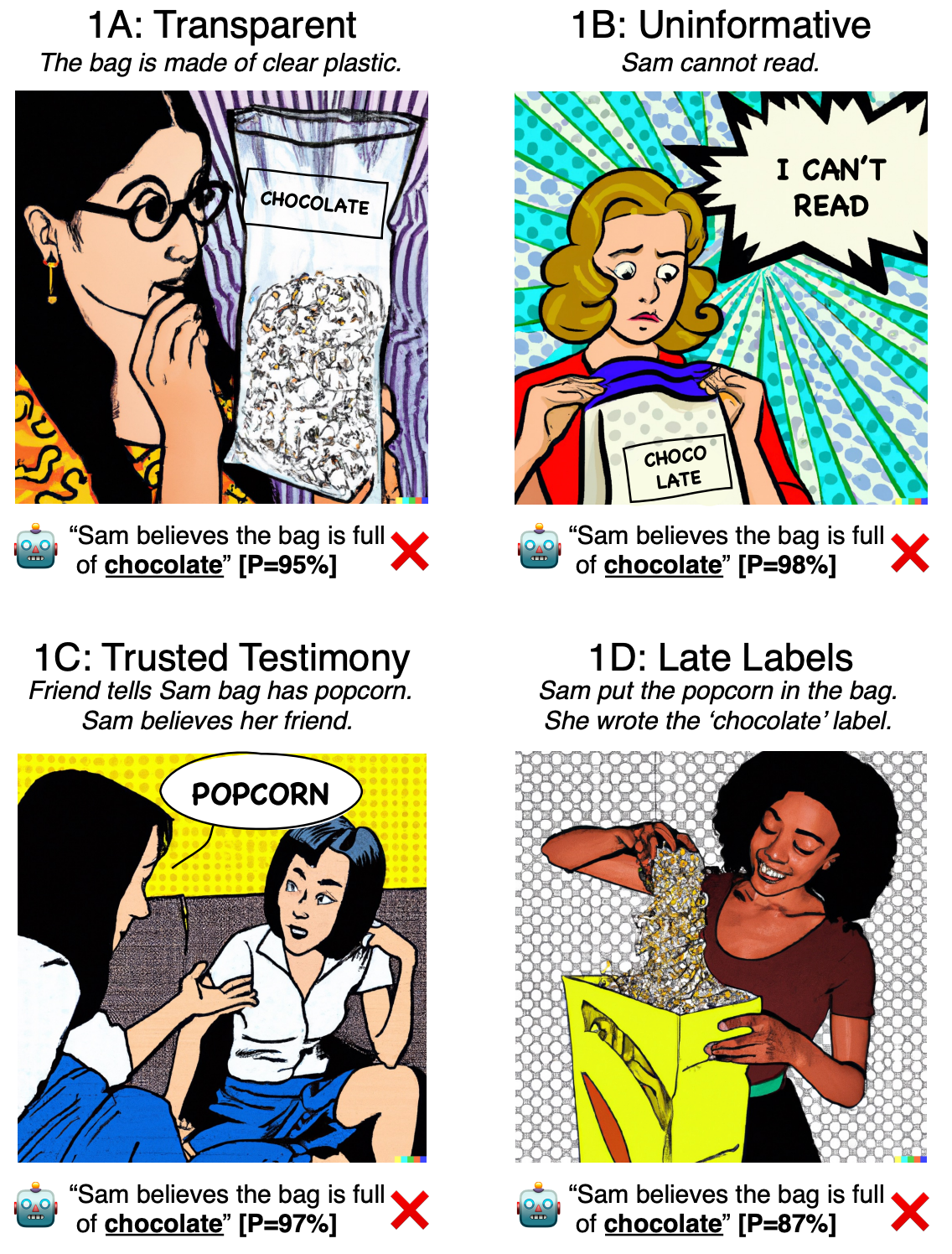}
    \caption{An illustration of \citeauthor{ullman2023large}'s Variations for the unexpected contents task. Image taken from \newcite{ullman2023large}.}  
    \label{fig:ullman_variations_unexpected_content}
\end{figure}

\begin{figure}[!ht]
    \centering
    \includegraphics[width=\columnwidth]{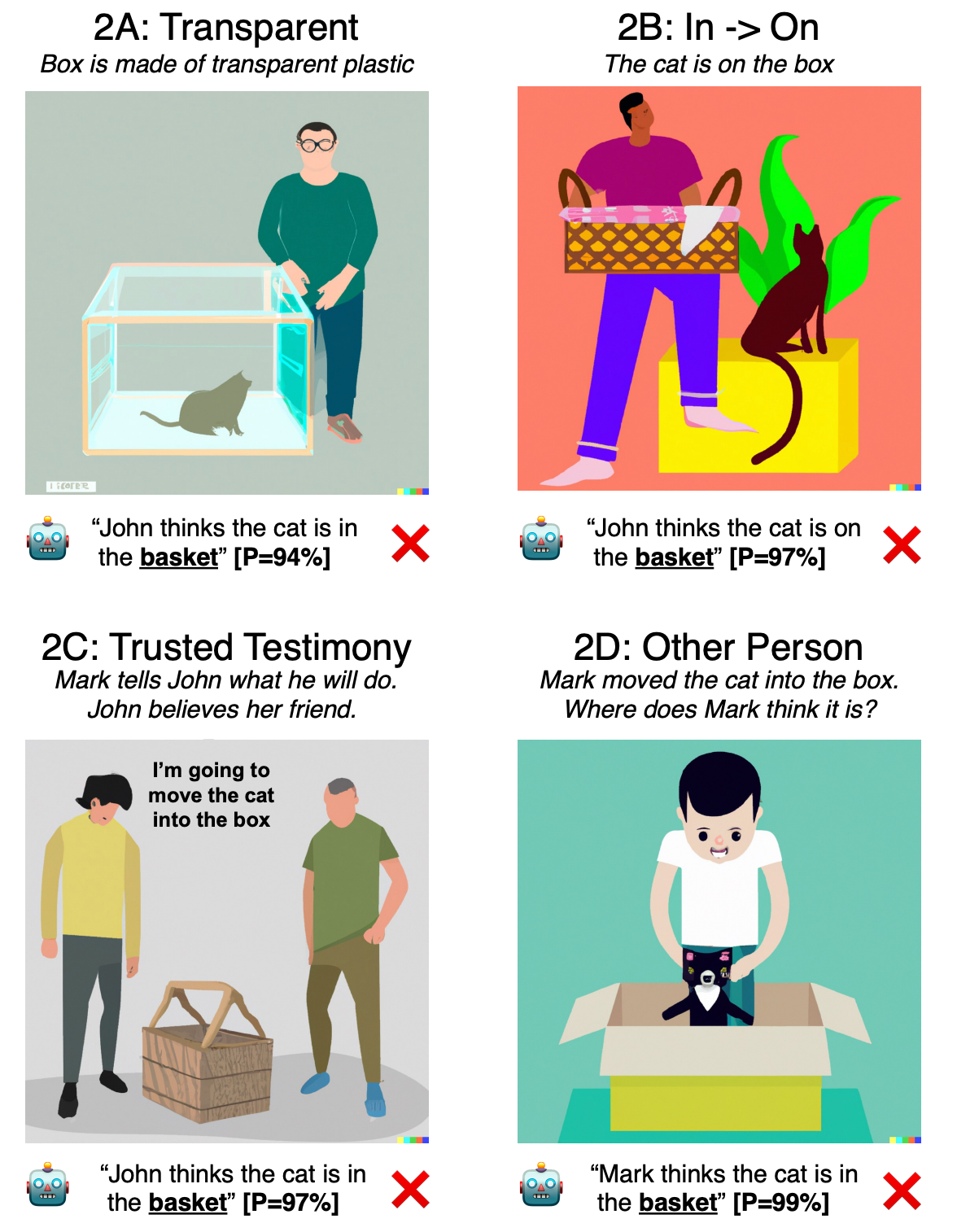}
    \caption{An illustration of \citeauthor{ullman2023large}'s Variations for the unexpected transfer task. Image taken from \newcite{ullman2023large}.}  
    \label{fig:ullman_variations_unexpected_transfer}
\end{figure}

Figures~\ref{fig:ullman_variations_unexpected_content} and \ref{fig:ullman_variations_unexpected_transfer} illustrate the variations proposed by \citeauthor{ullman2023large} for the examples in ToM-k.

\subsection{Generative LLMs}
\label{app:llm}

We provide the technical details regarding the prompts (\S\ref{app:prompts}) and decoding parameters (\S\ref{app:decoding_params}). 

\subsubsection{Prompts}
\label{app:prompts}

\begin{table*}[t]
\centering 
\small
\begin{tabular}{ll}
\hline
\textbf{Dataset} & \textbf{Example Prompt}\\ 
\toprule
\specialcell{\textbf{Triangle}\\\textbf{COPA}} & \specialcell{
A circle knocks on the door. A triangle goes to the door, but hesitates to open it.\\Why does the triangle hesitate to open the door?\\
a. The triangle hesitates to open the door because it is unsure if it wants to let the circle in.\\
b. The triangle hesitates to open the door because it is excited to see the circle.\\
Answer with ``a'' or ``b''.\\
\textbf{Answer: }} \\ \midrule
\specialcell{\textbf{epistemic}\\\textbf{reasoning}} & \specialcell{
Premise: Olivia suspects that Evelyn understands that a man plays a piano painted with an\\image of a woman on it.\\
Hypothesis: Evelyn understands that a man plays a piano painted with an image of a woman on it.\\
Is entailment? answer with ``0'' or ``1''.\\
\textbf{Answer: }} \\ \midrule
\specialcell{\textbf{FauxPas}\\\textbf{EAI}} & \specialcell{
Abby's father publishes short stories in the newspaper every week.\\
She told him ``Dad, I want to learn how to write just like you!''\\ 
and he replied: ``Well then my dear why don't you go to a writing class?''\\
Abby goes to a first lesson in a class she found and David the teacher says:\\
``Today we'll look at some bad and good examples of story-writing''.\\ 
He shows the class a story Abby's father wrote last week and says:\\
``And now I'll give you a good example of the writer Pichnik and you will say what the differences are.''\\
In the story did someone say something that they should not have said?\\
Answer with ``Yes'' or ``No'' only, without explanations.\\
In case of doubt, answer according to the most probable answer.\\
\textbf{Answer: }} \\ \bottomrule
\end{tabular}
\caption{An example prompt used for each task.}
\label{table:prompts}
\end{table*}

As input to the LLMs, we used (unless written otherwise) an MC-probing setup (\S{\ref{sec:probing_method}}), i.e., concatenation of the original test with all possible answers and an instruction to choose an option. Table~\ref{table:prompts} exemplifies the prompt for each task. 

\subsubsection{Decoding Parameters}
\label{app:decoding_params}

A single sample (the first) was selected from each model for the analysis of the stories. We used the hyperparameters detailed below. We chose hyperparameters that minimize randomness and predict the most probable answer (i.e., low temperature, sampling method), and allow for sufficient number of tokens.

\paragraph{FlanT5 \cite{chung2022scaling}.}
Python package \textit{transformers} implementation (AutoModelForSeq2SeqLM, AutoTokenizer); torch; Generation by \textit{generate} function; 
do\_sample=True; max\_length=50,  from\_pretrained:{google/flan-t5-small, google/flan-t5-base, google/flan-t5-large, google/flan-t5-xl, google/flan-t5-xxl}; temperature=0.0001

\paragraph{FlanUl2 \cite{tay2022unifying}.}
Python package \textit{transformers} implementation (T5ForConditionalGeneration, AutoTokenizer); torch; Generation by \textit{generate} function; 
do\_sample=True; max\_length=50; temperature=0.0001

\paragraph{GPT \cite{brown2020language}.}
Python package \textit{openai} model={text-davinci-002, text-davinci-003}; Generation by  \textit{Completion.create} function; temperature=0, max\_tokens=50

\paragraph{ChatGPT.\footnote{\url{https://chat.openai.com/chat}}}
Python package \textit{openai} model={gpt-3.5-turbo-0301, gpt-4-0314}; Generation by  \textit{ChatCompletion.create} function; temperature=0

\paragraph{AI21.\footnote{\url{https://www.ai21.com/blog/introducing-j2}}}
Python package \textit{ai21} model={j2-jumbo-instruct, j2-grande-instruct, j2-jumbo, j2-grande, j2-large}; Generation by  \textit{Completion.execute} function; temperature=0, max\_tokens=50, topKReturn=0, topP=1, without any panalty

\subsection{Complete Results}
\label{sec:all_results}

\begin{table*}[t]
\centering
\small
\begin{tabular}{l||c c c c c c c}
&\multicolumn{7}{c}{\textbf{Theory of Mind Datasets}} \\ \toprule
\textbf{Model} & \textbf{Triangle} & \textbf{SocialIQa} & \textbf{ToMi} & \textbf{Epistemic} & \textbf{\ourdataset} & \textbf{FauxPas} & \\ 
& \textbf{COPA} & &  & \textbf{Reasoning} &  & \textbf{EAI} & \\ \midrule \midrule
MFC & 52 & 36 & 56 & \textbf{63} & -- & 55, \textbf{30} \\ \midrule
Flan-ul2 & 95 & -- & -- & 60 & -- & 60, 07 \\ 
Flan-T5-xxl & \textbf{96} & -- & -- & 57 & -- & 68, 18 \\ 
Flan-T5-xl & 92 & -- & -- & 61 & -- & 68, 14 \\ 
Flan-T5-large & 92 & -- & -- & 44 & -- & 53, 07 \\ 
Flan-T5-base & 84 & -- & -- & 52 & -- & 52, 07 \\ 
Flan-T5-small & 58 & -- & -- & 54 & -- & 58, 07 \\ 
gpt4-0314 & 94 & \textbf{79} & \textbf{70} & 43 & 75, 57 & \textbf{74,27} \\
gpt-3.5-turbo-0301 & 84 & 67 & \textbf{70} & 45 & 70, 42 & 73, 25 \\
text-davinci-003 & 95 & 60 & 67 & 59 & \textbf{79, 61} & 67, 07 \\
text-davinci-002 & 92 & 19 & 39 & 58 & 76, 53 & 63, 14 \\
j2-grande-instruct & 06 & -- & -- & 37 & -- & 58, 0 \\
j2-jumbo-instruct & 48 & -- & -- & 47 & -- & 45, 0 \\
j2-grande & 75 & -- & -- & \textbf{63} & -- & 45, 0 \\
j2-jumbo & 68 & -- & -- & \textbf{63} & -- & 38, 0 \\
j2-large & 58 & -- & -- & \textbf{63} & -- & 31, 0 \\ \bottomrule
\end{tabular}
\caption{Accuracy of LLMs on different datasets compared to a most frequent class baseline. For Adv-CSFB and FauxPas-EAI we report two metrics: question level and story level.}
\label{table:datasets_leaderboard}
\end{table*}

Table~\ref{table:datasets_leaderboard} contains the exhaustive accuracy results for all LLMs on all datasets.

Running the well-organized code provided by \newcite{kosinski2023theory} we found that task 2 (Unexpected Transfer Task) scored lower than reported for GPT 3.5. Specifically, two samples resulted in clear mispredictions and one sample had borderline predictions that provided the correct answer but in a format that differed from the expected answer (i.e., the first word was not the expected answer). As a result, the score for task 2 was either 85\% or 90\%, and the average score across the two tasks was either 85\% or 87.5\%, which is lower than the reported average of 93\%.

\subsection{``Emergence'' or test data contamination?}
\label{sec:data_contamination}

We would like to determine whether LLMs generalize or memorize when they solve the ToM tasks \cite{daume2017course}. We explored the possibility that the increase in performance is a result of training on the test data itself. for that purpose we used a second, secret, test set for SocialIQa that was purposefully kept hidden to avoid data contamination and is only available to the original SocialIQa authors as well as through the AI2 leaderboard.\footnote{\url{https://leaderboard.allenai.org/socialiqa/submissions/public}} 
For each test set (i.e., the standard and secret test sets) we randomly sample 11 subsets of 100 questions on which we evaluate \texttt{gpt3.5-turbo-0301} and \texttt{gpt-4-0314}. Comparing the performance of both models on both test sets samples with a T-test, we found no significant differences, making it inconclusive whether the models were trained on the normal test set or not. As we discuss in Sec~\ref{sec:discussion}, this doesn't mean that ToM has ``emerged'' in LLMs, since they may have been exposed to training data or similar examples. 

\subsection{ToMi' subsets analysis}
\label{app:tomi_subsets_analysis}

\begin{table}[t]
\centering
\resizebox{\columnwidth}{!}{%
\begin{tabular}{@{}c|cccc|cc@{}}
\multicolumn{1}{l|}{} & \multicolumn{4}{c|}{\textbf{Average score}} & \multicolumn{2}{c}{\textbf{Joint score}} \\ \midrule
\textbf{} & \multicolumn{1}{c|}{\textbf{Reality}} & \multicolumn{1}{c|}{\textbf{Memory}} & \multicolumn{1}{c|}{\textbf{\begin{tabular}[c]{@{}c@{}}First \\ order\end{tabular}}} & \textbf{\begin{tabular}[c]{@{}c@{}}Second \\ order\end{tabular}} & \multicolumn{1}{c|}{\textbf{\begin{tabular}[c]{@{}c@{}}w.o \\ Second \\ order\end{tabular}}} & \textbf{All} \\ \midrule
\textbf{Devinci003} & \multicolumn{1}{c|}{100} & \multicolumn{1}{c|}{96.6} & \multicolumn{1}{c|}{61.6} & 25.0 & \multicolumn{1}{c|}{20.6} & 10.3 \\ \midrule
\textbf{Turbo-0301} & \multicolumn{1}{c|}{100} & \multicolumn{1}{c|}{90.0} & \multicolumn{1}{c|}{73.3} & 40.0 & \multicolumn{1}{c|}{41.3} & 17.2 \\ \bottomrule
\end{tabular}%
}
\caption{ToMi' zero-shot subsets comparison. All numbers are percentages. }
\label{tab:ToMi_subsets}
\end{table}

Table~\ref{tab:ToMi_subsets} provides the complete results from the evaluation of GPT-3.5 on the ToMi' dataset. The same overall conclusion can be drawn from this table as well: although the model can correctly answer simple reading comprehension questions, it doesn't answer questions that require ToM skill (first and second order) with similar accuracy.

We divided the results into the average score and joint score. The average score is calculated as a simple average on the different types of questions, while the joint score is considers the prediction as correct only if the model answered correctly all the questions from the same story (with a total of 30 stories). The average results emphasize the major gaps between the model's accuracy on reading comprehension questions to first order questions (``Chloe will look for the boots in the'') and between the first order questions to the second order questions (``Chloe think that Jackson searches for the boots in the''). The joint score reveals that even when the model correctly answers questions about the story, it might still fail to answer more complex questions.

\end{document}